\pdfoutput=1

\documentclass[11pt]{article}

\usepackage[]{emnlp2022}

\usepackage{times}
\usepackage{latexsym}

\usepackage[T1]{fontenc}

\usepackage[utf8]{inputenc}

\usepackage{microtype}

\usepackage{inconsolata}
\usepackage{graphicx}
\usepackage{subcaption}
\usepackage{enumitem}
\setlist[itemize]{noitemsep, topsep=0pt}
\usepackage{amsmath}
\usepackage{xcolor}
\usepackage{colortbl}

\newcommand{\highlightred}[1]{%
  \colorbox{red!30}{$\displaystyle#1$}}
\newcommand{\highlightblue}[1]{%
  \colorbox{blue!40}{$\displaystyle#1$}}
\newcommand{\highlightyellow}[1]{%
  \colorbox{yellow!50}{$\displaystyle#1$}}
\newcommand{\highlightorange}[1]{%
  \colorbox{orange!50}{$\displaystyle#1$}}
\newcommand{\highlightgreen}[1]{%
  \colorbox{green!30}{$\displaystyle#1$}}
\newcommand{\highlightcyan}[1]{%
  \colorbox{cyan!30}{$\displaystyle#1$}}
\newcommand{\highlightmagenta}[1]{%
  \colorbox{magenta!30}{$\displaystyle#1$}}
  
\everypar{\looseness=-1}

\title{Extractive Summarization of Legal Decisions using Multi-task Learning and  Maximal Marginal Relevance}

\author{
    Abhishek Agarwal \and Shanshan Xu \and Matthias Grabmair\\
    Technical University of Munich,  Germany\\
    \texttt{\{abhishek.agarwal, shanshan.xu, matthias.grabmair\}@tum.de} \\
    }

\begin{document}
\maketitle
\begin{abstract}
Summarizing legal decisions requires the expertise of law practitioners, which is both time- and cost-intensive. This paper presents techniques for extractive summarization of legal decisions in a low-resource setting using limited expert annotated data. We test a set of models that locate relevant content using a sequential model and tackle redundancy by leveraging maximal marginal relevance to compose summaries. We also demonstrate an implicit approach to help train our proposed models generate more informative summaries. Our multi-task learning model variant leverages rhetorical role identification as an auxiliary task to further improve the summarizer. We perform extensive experiments on datasets containing legal decisions from the US Board of Veterans' Appeals and conduct quantitative and expert-ranked evaluations of our models. Our results show that the proposed approaches can achieve ROUGE scores vis-à-vis expert extracted summaries that match those achieved by inter-annotator comparison.
\end{abstract}

\section{Introduction}

In common-law systems, law practitioners research large numbers of legal decisions from past cases to find similar precedents that justify their arguments and lead to favorable outcomes. The analysis can be time-consuming and expensive as these documents are long and verbose, and understanding them requires legal expertise. Automatic summarization of legal documents can help expedite the process cost-effectively. However, the limited availability of expert-annotated summaries makes it challenging to design such automated systems to assist paralegals, lawyers, and other law practitioners.

Extractive summarization aims to identify and extract essential sentences from the source document to compose the corresponding summary. It is more common in the legal domain due to the complexity of the legal language and the scarcity of labeled data. By contrast, abstractive summarization generates an abstract representation that captures the salient ideas of the source text and might contain new words and phrases not present in the source document.

One of the main challenges of extractive summarization is the redundancy in legal documents, as legal decisions can often contain several semantically similar sentences. Our objective is to generate summaries that provide maximum information while minimizing redundancy. Maximal Marginal Relevance (\textit{MMR}) \cite{carbonell1998use} has proved to be an effective tool to tackle redundancy explicitly \cite{zhong2019automatic} by balancing the importance of query relevance and diversity. However, more recent methods like \textit{MMR-Select} can use neural models as a substitute for query relevance. Additionally, we can train the neural models to handle the redundancy implicitly by adding a redundancy loss term \cite{xiao2020systematically}.

Another challenge is the low availability of expert annotated summarization datasets in the legal domain. In this work, we leverage large amounts of unlabeled data along with the small annotated datasets to gain maximum performance. Pre-trained transformers like BERT \cite{devlin2019bert} can improve the performance of downstream tasks, such as summarization, even with limited labeled data. However, such models trained on the general domain may fail to capture the intricacies of the domain-specific vocabulary used in legal decisions. The domain-specific variants of BERT \cite{chalkidis2020legal, zheng2021does} pre-trained on large corpora of legal texts can help better embed the legal terms and achieve robust performance in various legal-specific downstream tasks like argument mining \cite{xu2021accounting},  rhetorical role labeling \cite{bhattacharya2021deeprhole}, and legal citation recommendation \cite{huang2021context}.

To maximize the summarization performance, we also leverage Multi-task Learning (MTL) by aggregating training samples from several smaller datasets of multiple related tasks. MTL helps the model learn shared representations between the primary task (summarization) and the auxiliary task (rhetorical  role identification) to generalize better. The identification of rhetorical roles involves identifying the function of different sentences to understand underlying reasoning and argument patterns in legal decisions. Previous works have often used rhetorical role labeling as a precursor to extractive summarization to improve performance \cite{zhong2019automatic, bhattacharya2021incorporating}. In this paper, we explore the idea of using rhetorical role identification as an auxiliary task to augment our annotated dataset and help generate better summaries.

In brief, we consider our contributions to the extractive summarization of legal documents as follows:
\begin{itemize}
    \item We generate informative summaries with maximum information and minimum redundancy in a low-resource setting. Our experiments demonstrate a general improvement in ROUGE scores for the proposed approaches.
    
    \item We further improve the summarizer using a multi-task setting by combining extractive summarization and rhetorical role labeling. The quantitative evaluation demonstrates that the multi-task models perform better than the single-task models.

    \item We evaluate the generated summaries qualitatively with the help of a legal expert. In contrast to the quantitative evaluation, the qualitative results show that our proposed approaches rank at least as good as human annotators.\footnote{Our code is available  \href{https://github.com/TUMLegalTech/summarization\_emnlp22}{here}}

\end{itemize}

\section{Related Work}\label{section:related_work}

\subsection{Extractive Summarization}\label{subsection:related_work_extractive_summarization_legal}

\citet{galgani2012combining} developed a rule-based approach to summarization that uses a knowledge base, statistical information, and other handcrafted features like POS tags, specific legal terms, and citations. \citet{kim2012summarization} propose a graph-based summarization system that constructs a directed graph for each document where nodes are assigned weights based on how likely words in a given sentence appear in the conclusion of judgments. CaseSummarizer \cite{polsley2016casesummarizer}, an automated text summarization tool, uses word frequency augmented with additional domain-specific knowledge to score the sentences in the case document.  \citet{liu2019extracting} propose a classification-based approach that uses several handcrafted features as input. However, such techniques require knowledge engineering of different features and do not tackle redundancy in legal decisions. Recently, various proposed approaches have tried to address redundancy in legal decisions for purposes of summarization. \citet{zhong2019automatic} hypothesize that the iterative selection of predictive sentences using a CNN-based train-attribute-mask pipeline followed by a Random Forest classifier to distinguish between sentences containing Reasoning/EvidentialSupport and other types. \textit{MMR} then selects the final sentences for the summary. \cite{bhattacharya2021incorporating} demonstrate an unsupervised approach named DELSumm that generates extractive summaries by incorporating guidelines from legal experts into an optimization problem that maximizes the informativeness and content words, as well as conciseness. In this work, we use an \textit{MMR}-based variant which tackles redundancy explicitly and can be combined with a neural classifier to generate summaries. It alleviates the need to engineer handcrafted features or specific expert guidelines to prevent redundancy. 

\subsection{Rhetorical Role Labeling}\label{subsection:related_work_rhetorical_labeling}

\citet{saravanan2010identification} propose a rule-based system along with a Conditional Random Field (CRF) approach to identify the different segments. \citet{nejadgholi2017semi} proposed a semi-supervised approach to searching legal facts in immigration-specific case documents by using an unsupervised word embedding model to aid the training of a supervised fact-detecting classifier using a small set of annotated sentences. The authors in \citep{walker2019automatic} compare the performance between rule-based scripts and ML algorithms to classify sentences that state findings of fact. \citet{bhattacharya2019identification} explore the use of hierarchical BiLSTM models by adding an attention layer and experiment with the pre-trained word and sentence embeddings \cite{bhattacharya2021deeprhole}.  \cite{savelka2021lex} annotated legal cases from seven countries in six languages using a structural type system and found that Bi-GRU models could be generalized for data across different jurisdictions to some degree. Despite copious work, there are very few annotated rhetorical role datasets in the legal domain. In this work, we use rhetorical role labeling as an auxiliary task to augment our annotated dataset and help generate better summaries.

\section{Data}\label{section:dataset}

We use the dataset containing single-issue Post-Traumatic Stress Disorder decisions from the US Board of Veterans' Appeals\footnote{\url{https://www.bva.va.gov}} (BVA) by \cite{zhong2019automatic}. These cases focus on veterans' appeals for benefits for a PTSD disability connected to stressful experiences during military service. The dataset is a sample from the BVA database that has been constrained to single-issue cases focusing on PTSD. In the texts, the BVA reviews the available evidence and either makes a finding that it warrants an award for service-connected PTSD (granted) or not (denied), or refers the case back to a lower administrative division for further development (remand). The dataset consists of 112 decisions and the corresponding expert annotated gold-standard summaries. We have 92 cases (48 remanded, 28 denied, 16 granted) in the training set with one annotated summary each. Another 20 cases (10 remanded, 6 denied,  4 granted) constitute the test set, for which there are four extractive summaries by different annotators and two drafted abstractive summaries. Each annotator chose a 6-10 sentence long summary based on predefined guidelines, out of which they selected 1-3 sentences each from the \textit{Reasoning} and \textit{Evidence} annotation type. The \textit{Reasoning} sentences connect the outcome to the facts, while \textit{Evidence} sentences add more information to support the former.

For rhetorical role labeling, we use two different datasets containing 50 plus 25 annotated BVA decisions\footnote{\url{https://github.com/LLTLab/VetClaims-JSON}}\citep{walker2019automatic}. The decisions in the larger dataset have partial annotations, so we keep only decisions that have annotations for at least 60\%\footnote{The remaining sentences with missing annotations contain sentences from annotation types other than \textit{Evidence} or \textit{Reasoning}, so we automatically annotate these sentences with the \textit{Others} type.} of the sentences in each decision. It results in 28 decisions consisting of 17 denied and 11 granted outcomes, while the smaller dataset contains 10 denied and 15 granted decisions. We map the different annotation types of the two datasets to a uniform type system of six annotation types. We merged the different annotation types for our experiments, resulting in 1889 \textit{Evidence/Reasoning} and 3728 \textit{Others} sentences. Therefore, the final dataset has 53 decisions with 7473 binary sentence-level annotations.

The datasets are from different time periods; therefore, the decisions have a slightly different document structure. We remove the meta-information like the case number, dates, judge names, names of the witnesses, and other similar information to have a more uniform layout. We keep only the following sections (if present) from each decision in the dataset:
\begin{itemize}
    \item \textit{Order}
    \item \textit{Finding of Fact}
    \item \textit{Conclusion of Law}
    \item \textit{Reasons and Bases for Finding and Conclusion}
    \item \textit{Remand}
    \item \textit{Reasons for Remand}
\end{itemize}

Additionally, we use the SpaCy\footnote{\url{https://spacy.io}} pipeline enhanced with additional handcrafted rules to segment the sentences in each document for the summarization dataset.\footnote{To validate the performance of the sentence segmentation, we manually segment 11 decisions (6 remanded, 3 denied, 2 granted) and compare the matches. In terms of Recall, our approach and the legal text sentence segmenter \citep{savelka2017sentence} score \textbf{0.937} and \textbf{0.905}, respectively.} After pre-processing, the average number of sentences per decision in the summarization and rhetorical role labeling datasets is \textbf{77.29 (± 52.28}) and \textbf{118.37 (± 78.33)}, respectively.

\section{Our Approach}\label{section:our_approach}

\subsection{Sentence Embeddings}\label{section:sentence_embeddings}

Sentence embeddings map an input sentence to a fixed-size dense vector representation. SentenceBERT \cite{reimers-gurevych-2019-sentence} has recently emerged as an effective tool to derive semantically meaningful sentence embeddings. However, the lack of a domain-specific labeled entailment dataset required to train it makes it inaccessible for us. Alternatively, we can use a BERT model to extract a sentence embedding by pooling the embeddings for each token in the sentence. The mean-pooling operation outputs a 768-dimensional fixed-sized representation for each sentence. Such embeddings generalize quite well and provide a good starting point for training our sequential models later. This work uses the legal-domain specific transformer Legal-BERT \citep{zheng2021does} trained on the Harvard Law case corpus' 3,446,187 legal decisions to generate the sentence embeddings.

\subsection{Weighted Loss Function}\label{section:weighted_loss_function}

The conventional cross-entropy loss function for the extractive summarization results in poor classification performance due to the class imbalance. For each decision, on average, we have very few positive labels (5-6 sentences) for each summary, which results in a highly imbalanced dataset. To tackle this issue, we use the weighted cross-entropy loss function that puts more emphasis on positive labels by manually rescaling the weights for each class.

\begin{gather*}
    \highlightgreen{w_{c}} = \frac{\#samples }{\#classes \times \#samples_c}\\
    L_{CE} = -\sum_{c=1}^M \highlightgreen{w_{c}} \highlightorange{(y_{o,c}\log(p_{o,c}))}
\end{gather*}

\subsection{Maximal Marginal Relevance}\label{section:maximal_marginal_relevance}
Maximal Marginal Relevance (\textit{MMR}) \cite{carbonell1998use} iteratively (greedily) selects sentences for the summary while balancing the query relevance and diversity:

\begin{multline*}
\operatorname{MMR} = arg \max\limits_{s_{i}\epsilon D \backslash \hat{S}} [\highlightyellow{\lambda} \highlightred{Sim(s_{i}, Q)}\\
- \highlightyellow{(1-\lambda)} \max\limits_{s_{j} \epsilon \hat{S}} \highlightblue{Sim(s_{i}, s_{j})}]
\end{multline*}

The parameter $\lambda$ helps control the redundancy (novelty) in the extracted summary. We use cosine similarity to calculate the similarity between two sentence embeddings. The query $Q$ represents the case document by taking the average embeddings of all the sentences in the decision. \citet{xiao2020systematically} propose \textit{MMR-Select} as an alternative approach to eliminate redundancy explicitly. It eliminates the greedy method, computing the query relevance to find suitable candidates with a neural model. \textit{\textit{MMR}} is more robust as it picks candidate sentences using the confidence scores, $P(y_{i})$, produced by the neural model.

\begin{multline*}
\operatorname{MMR-Select} = arg \max\limits_{s_{i}\epsilon D \backslash \hat{S}} [\highlightyellow{\lambda} \highlightred{P(y_{i})}\\
- \highlightyellow{(1-\lambda)} \max\limits_{s_{j} \epsilon \hat{S}} \highlightblue{Sim(s_{i}, s_{j})}]
\end{multline*}

\subsection{Redundancy Loss}\label{section:redundancy_loss}

The major limitation of explicit methods like \textit{MMR} is the disconnect between the sentence scoring and sentence selection phases. Such techniques rely on the classifier to score the sentences in the document and check for redundancy later when selecting the final sentences for the summary. Thus, the classifier used to generate the confidence score does not implicitly learn how to handle redundancy. We can generate more informative summaries by teaching the neural model to avoid picking similar sentences. \citet{xiao2020systematically} propose adding a redundancy loss term $L_{RD}$ to the cross-entropy loss function that penalizes the model for choosing two similar sentences with high confidence scores. The parameter $\beta$ balances the importance we assign to the $L_{CE}$ and $L_{RD}$. The neural models tend to classify more sentences as part of the summary for longer case documents. Therefore, we scale the redundancy loss $L_{RD}$ defined by \citet{xiao2020systematically} to ensure that it does not explode as the length of the document increases, preventing it from overshadowing the cross-entropy loss.

\begin{gather*}
L = \highlightcyan{\beta}L_{CE} + \highlightcyan{(1-\beta)}L_{RD}\\
L_{RD} = \highlightmagenta{\frac{1}{n^{2}}} \sum_{i=1}^{n}\sum_{j=1}^{n}\highlightred{P(y_{i})}\highlightred{P(y_{j})}\highlightblue{Sim(s_{i}, s_{j})}
\end{gather*}

\subsection{Extractive Summarization}\label{section:extractive_summarization}
\subsubsection{Single-Task Models}

\begin{figure}[htbp]
    \centering
    \includegraphics[width=1\linewidth]{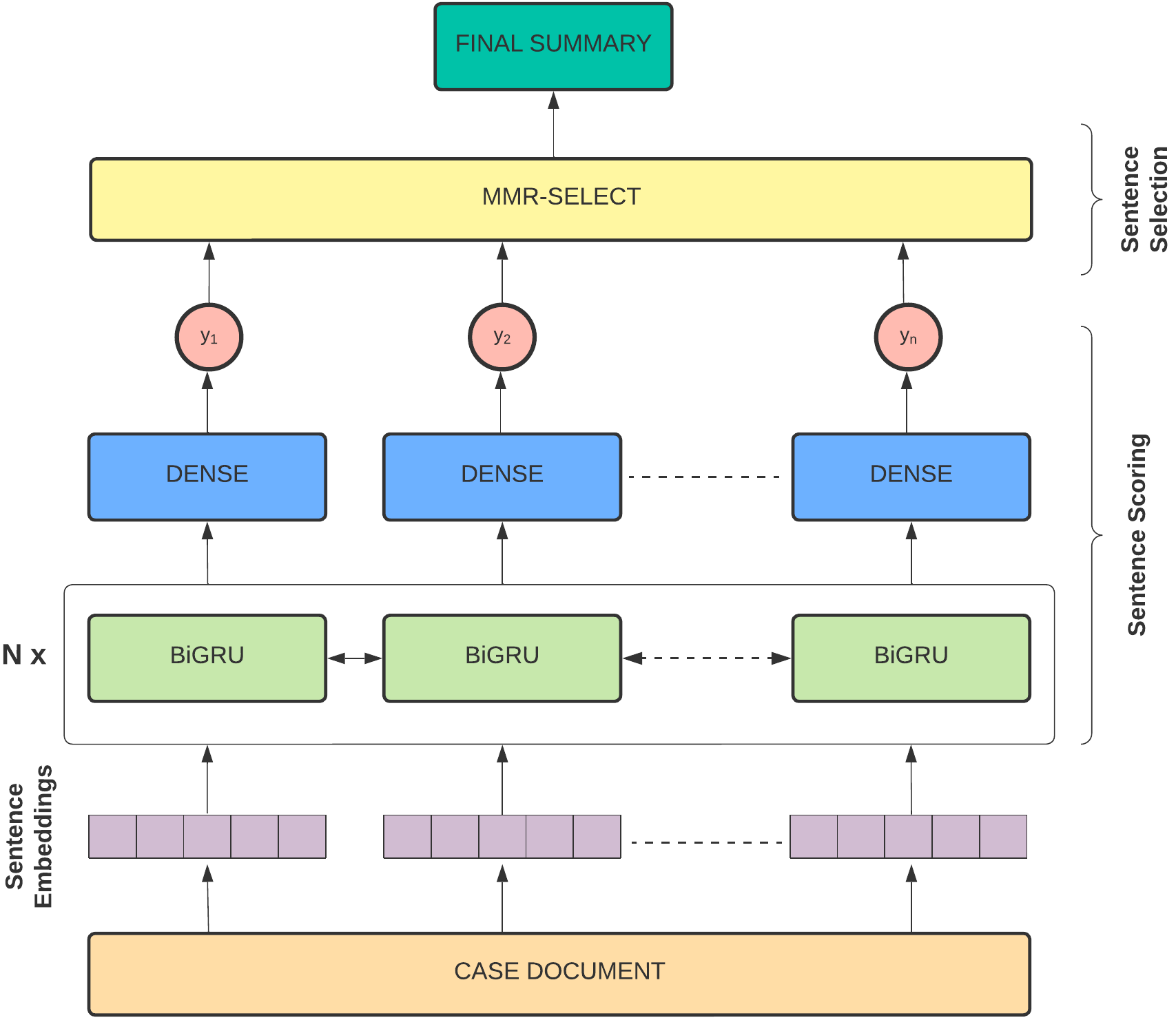}
    \caption{Architecture of the proposed Single-Task Extractive Summarization models: ST and ST+RdLoss}
    \label{fig:st_architecture}
\end{figure}

We define extractive summarization as a binary classification problem where the proposed models decide whether a given sentence belongs to the fixed-length summary or not. We use the proposed models to generate the summary only for \textit{Reasoning/Evidence} sentences as we can extract perfect matches for the other rhetorical role sentences (e.g., case issue and procedural background) by using regular expressions\footnote{This is particular to the task of summarizing BVA decisions as introduced by \cite{zhong2019automatic}}. Our proposed models consist of two phases: Sentence Scoring and Sentence Selection, as shown in Figure \ref{fig:st_architecture}. Initially, we use the approach explained in Section \ref{section:sentence_embeddings} to generate the embeddings for all the sentences in a given case document.

\textbf{Sentence Scoring}: Bidirectional Gated Recurrent Units (Bi-GRU) use two GRUs to simultaneously encode the sentence embeddings in both forward and backward directions. The concatenation of the forward and backward hidden states gives us the representation of each input sequence. A fully connected dense layer followed by the non-linear softmax layer predicts the probability distribution over the two classes. We do not use transformer-based architectures for various reasons, including limited training data, input size limitations, and required computational resources.

\textbf{Sentence Selection}: \textit{MMR}, discussed in Section \ref{section:maximal_marginal_relevance}, uses the sentence embeddings and confidence scores generated by the neural model to tackle redundancy explicitly and select the final sentences for the summary.

Accordingly, we refer to our single-task model described above as \textbf{ST}. We also propose another model, \textbf{ST+RdLoss}, to implicitly handle redundancy by using Redundancy Loss.

\begin{figure}[htbp]
    \begin{subfigure}{1\linewidth}
      \centering
      \includegraphics[width=1\linewidth]{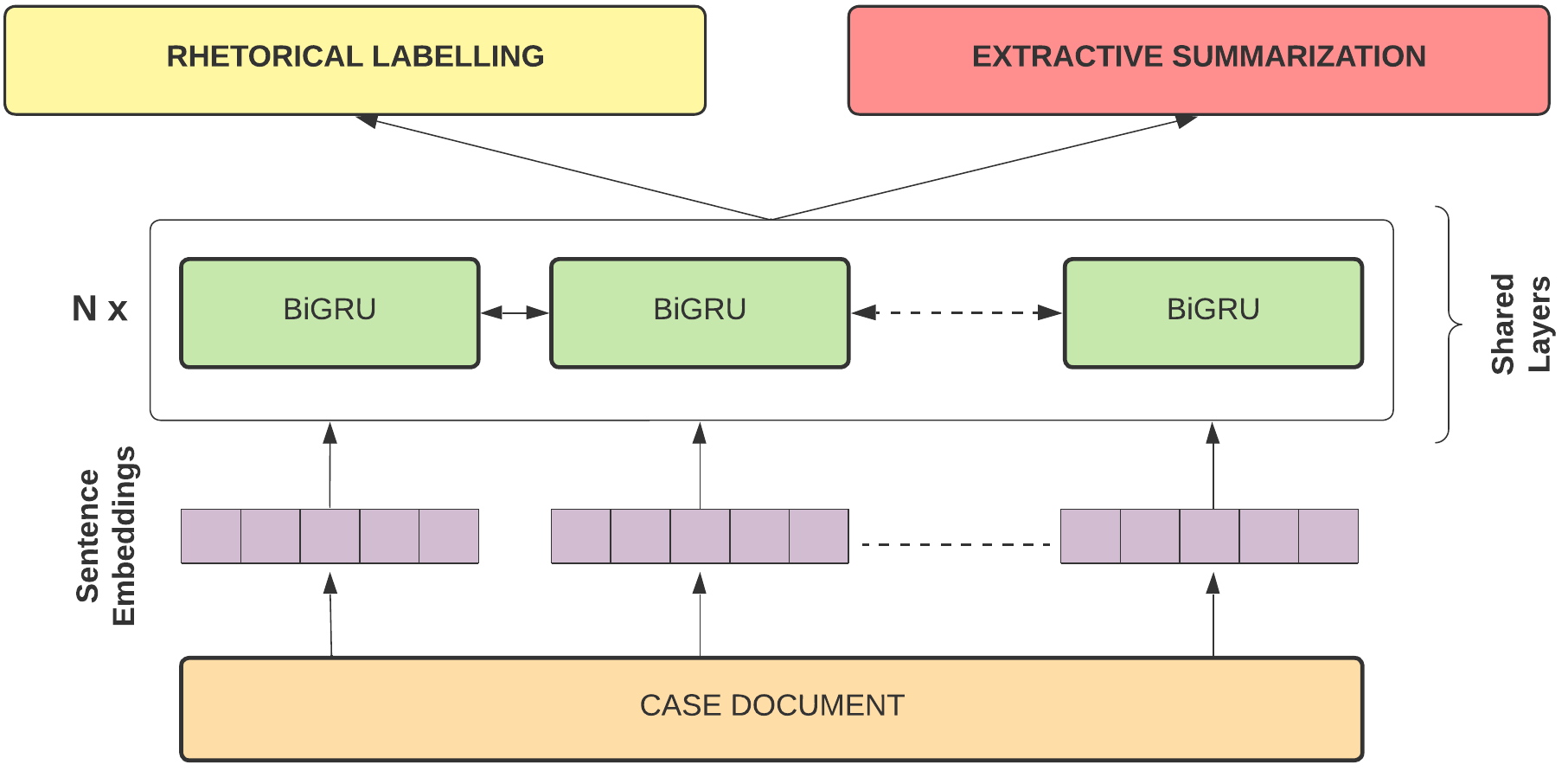}
      \caption{Shared Multi-Task Model}
      \label{fig:mtl_shared}
    \end{subfigure}
    
    \bigskip
    
    \begin{subfigure}{0.96\linewidth}
      \centering
      \includegraphics[width=1\linewidth]{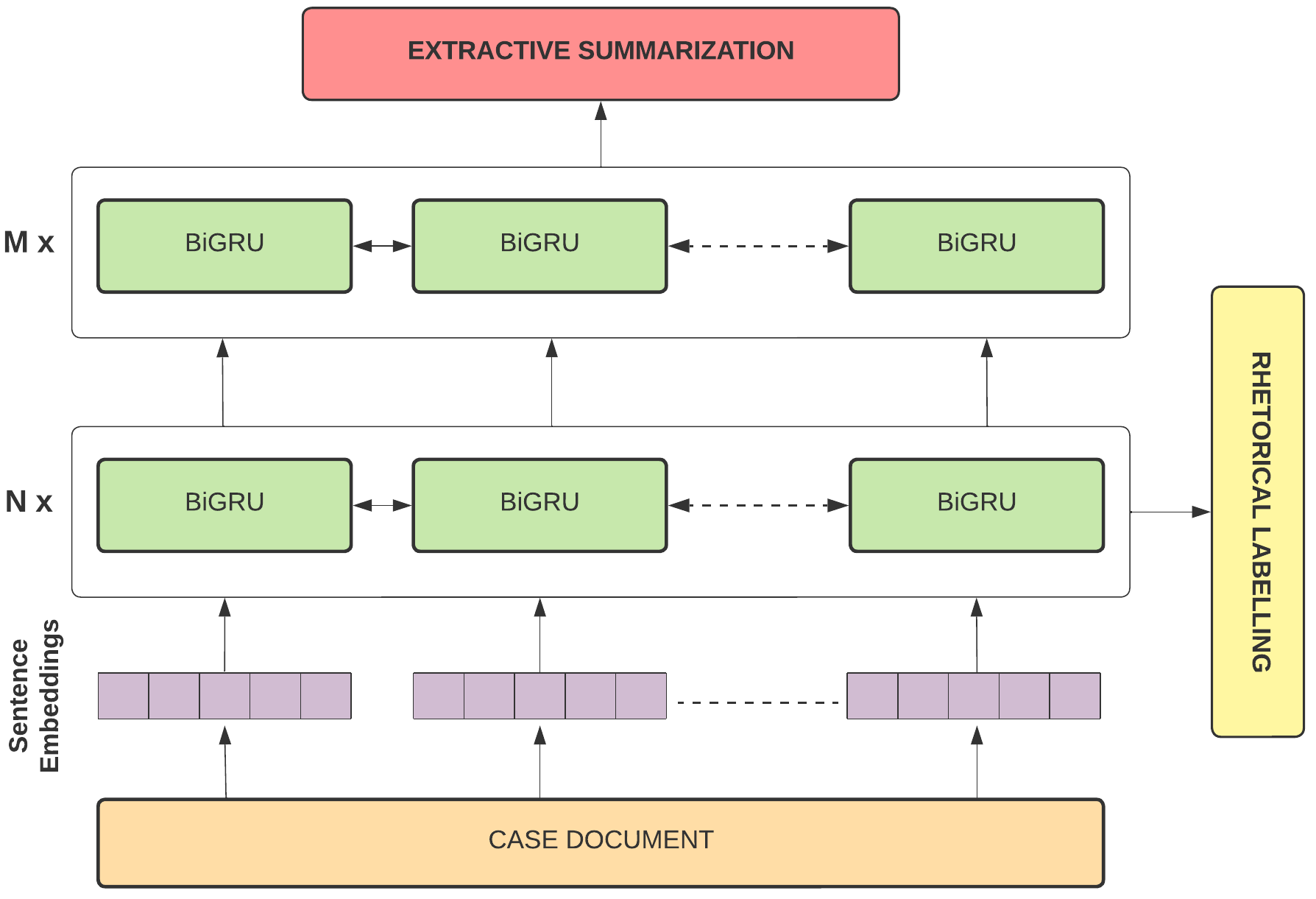}
      \caption{Hierarchical Multi-Task Model}
      \label{fig:mtl_hierarchical}
    \end{subfigure}
    \caption{Architecture of the proposed Multi-Task models: MT-Shared and MT-Hierarchical, respectively}
    \label{fig:mtl_architecture}
\end{figure}

\subsubsection{Multi-Task Models}

Learning multiple tasks by jointly optimizing more than one criterion can help leverage the correlation between related tasks to improve performance \cite{liu2016recurrent, ruder2017overview, elnaggar2018multi}. We propose using rhetorical role labeling as an auxiliary task to benefit the primary task of extractive summarization. We consider rhetorical role labeling a binary classification problem where the trained model learns to distinguish between \textit{Reasoning/Evidence} and other rhetorical roles. This objective is similar to extractive summarization, where we only include sentences from \textit{Reasoning/Evidence} in the summary.

Accordingly, we propose two multi-task models: \textbf{MT-Shared} and \textbf{MT-Hierarchical}, as shown in Figure \ref{fig:mtl_architecture}. The first model shares the same Bi-GRU layers for both tasks. The second model follows a hierarchical order where rhetorical role labeling only uses lower-level Bi-GRU layers. The extractive summarization shares the lower-level Bi-GRU label with rhetorical labeling and has additional Bi-GRU layers on top to learn different features. Besides the shared Bi-GRU layers, each task has its task-specific layer comprising a fully connected dense layer followed by the non-linear softmax layer to predict the probability distribution over the two classes. Additionally, \textbf{MT-Shared+RdLoss} and \textbf{MT-Hierarchical+RdLoss} use Redundancy Loss discussed in Section \ref{section:redundancy_loss}.

\section{Experiments}

\subsection{Baseline and Comparison}

We evaluate\footnote{We do not compare the results with previous work by \citet{zhong2019automatic} due to the difference in the structure of the summaries generated and reproducibility issues.} the performance of our proposed approaches with several baseline methods commonly used for extractive summarization. The two most common unsupervised methods are \textit{MMR} and TextRank \citep{mihalcea2004textrank}. TextRank uses a graph-based ranking system similar to PageRank to find relevant sentences. Since we are interested in keeping only Reasoning/Evidence sentences for the summary, we also employ a binary classifier to filter out such sentences from the other rhetorical roles. We use CatBoost \textit{(CB)} \cite{dorogush2018catboost} and a GRU sequence labeler as the binary classifiers to identify the Reasoning/Evidence sentences \footnote{Our CatBoost model classifies one sentence at a time irrespective of the case document, whereas the GRU model takes all the sentences in the document as the input achieving an F1 score of \textbf{0.917} and \textbf{0.914}, respectively.}. The unsupervised methods then use the filtered-out sentences as the input to extract summaries. For \textit{MMR}, we use the sentence embeddings discussed in Section 4.1 as the input, and cosine similarity measures the similarity between sentences. We generate the summaries for the \textit{TextRank} approach using the Gensim\footnote{https://radimrehurek.com/gensim/} package that uses \textit{BM25} scoring instead of TF-IDF or cosine similarity \citep{barrios2016variations}. Thus we have four different baseline methods: \textbf{RL-CB+MMR} \textit{(Cosine)}, \textbf{RL-GRU+MMR} \textit{(Cosine)}, \textbf{RL-CB+TextRank} \textit{(BM25)} and \textbf{RL-GRU+TextRank} \textit{(BM25)}.

\subsection{Implementation Details}
We use five-fold cross-validation to find the best hyperparameters for our baseline and proposed models. The parameter $\lambda$ for our \textit{MMR}-based baseline models is determined using the training set described in Appendix \ref{appendix:baseline}. We use hyperparameter tuning for \textit{MMR} and Redundancy Loss to find the best values for $\lambda$ and $\beta$, respectively. To train the multi-task learning models, we use the conventional cross-entropy loss function for the rhetorical role labeling task and alternate between the two datasets after every iteration. We randomly oversample the rhetorical role labeling task data to match the number of samples in the extractive summarization dataset. Therefore, in each mini-batch, we have training samples from either the extractive summarization or the rhetorical role labeling dataset and switch between them every other batch. We report the best set of hyperparameters corresponding to each model in Appendix \ref{appendix:proposed_models}.

\subsection{Evaluation Metrics}
We use Recall to measure the performance for the binary classification problem. It measures how many sentences selected by the model are also part of the expert annotated summary. The Recall-Oriented Understudy for Gisting Evaluation (ROUGE) \cite{lin2004rouge} score counts the number of overlapping word sequences between candidate and reference summaries. ROUGE-1 and ROUGE-2 measure the unigram and bigram overlap, respectively. ROUGE-L finds the longest common subsequence matches to reflect sentence-level word ordering better.

\section{Results and Analysis}

\subsection{Quantitative Evaluation}
In Table \ref{tab:summary_overview}, we report the number of sentences and tokens for the summaries in the test set generated by the different approaches. Our baseline and proposed models, which use MMR or MMR-Select, take the number of sentences as the input to generate the final summaries. Since the average number of sentences varies from 4.5 to 5.85 for the expert annotators, we choose n=5 as the suitable value for our models. For TextRank models, we require the number of tokens as the input, which we set to be 160 based on the statistics of the training set. Additionally, we observe that Annotator 2 tends to pick more sentences than the rest of the annotators resulting in the highest average token count of 171.3, while Annotator 1 and Annotator 3 pick fewer sentences on average. Our models' summaries have token counts similar to that of Annotator 4.

\begin{table*}[tb]
\centering
\small
\begin{tabular}{lc|c|c|c}
\hline
\multicolumn{1}{c}{}   & \multicolumn{1}{c|}{Sentences} & \multicolumn{1}{c|}{Tokens} & \multicolumn{1}{l|}{$\lambda$} & \multicolumn{1}{l}{$\beta$} \\ \hline \hline
Annotator 1            & 4.6 ± 1.46                     & 130.55 ± 51.11              & -                           & -                        \\
Annotator 2            & 5.85 ± 0.67                    & 171.3 ± 44.43               & -                           & -                        \\
Annotator 3            & 4.5 ± 0.889                    & 137.5 ± 37.99               & -                           & -                        \\
Annotator 4            & 5.55 ± 2.23                    & 149.85 ± 79.44              & -                           & -                        \\ \hline \hline
RL-CB+MMR              & 5 ± 0                          & 151.55 ± 33.76              & 0.9                         & -                        \\
RL-CB+TextRank         & 5.45 ± 1.31                    & 153.2 ± 9.57                & -                           & -                        \\ \hline
RL-GRU+MMR             & 5 ± 0                          & 156.6 ± 38.81               & 0.9                         & -                        \\
RL-GRU+TextRank        & 5.05 ± 1.23                    & 158.7 ± 13.24               & -                           & -                        \\ \hline \hline
ST                     & 5 ± 0                          & 156.2 ± 35.48               & 0.8                         & -                        \\
ST+RdLoss              & 5 ± 0                          & 149.35 ± 33.92              & 0.9                         & 0.85                     \\ \hline
MT-Shared              & 5 ± 0                          & 150.8 ± 27.2                & 0.6                         & -                        \\
MT-Shared+RdLoss       & 5 ± 0                          & 157 ± 34.1                & 0.9                         & 0.775                    \\
MT-Hierarchical        & 5 ± 0                          & 151.75 ± 28.04              & 0.9                         & -                        \\
MT-Hierarchical+RdLoss & 5 ± 0                          & 154.75 ± 36.42              & 1                           & 0.9                      \\ \hline
\end{tabular}
\caption{Overview of the \textbf{number of sentences} and \textbf{tokens} in the summaries generated by expert annotators and models. Values of parameters, \textbf{$\lambda$} and \textbf{$\beta$}, for \textit{MMR}, \textit{MMR} and RdLoss-based approaches.}
\label{tab:summary_overview}
\end{table*}

\begin{table*}[htbp]
\centering
\small
\begin{tabular}{lrrr|rrr|rrr|rrr}
\hline
                       & \multicolumn{3}{c|}{Annotator 1}                                                        & \multicolumn{3}{c|}{Annotator 2}                                                        & \multicolumn{3}{c|}{Annotator 3}                                                        & \multicolumn{3}{c}{Annotator 4}                                                        \\ \hline
                       & \multicolumn{1}{r}{R-1}    & \multicolumn{1}{r}{R-2}    & \multicolumn{1}{r|}{R-L}    & \multicolumn{1}{r}{R-1}    & \multicolumn{1}{r}{R-2}    & \multicolumn{1}{r|}{R-L}    & \multicolumn{1}{r}{R-1}    & \multicolumn{1}{r}{R-2}    & \multicolumn{1}{r|}{R-L}    & \multicolumn{1}{r}{R-1}    & \multicolumn{1}{r}{R-2}    & \multicolumn{1}{r}{R-L}     \\ \hline \hline
Annotator 1            & 100                         & 100                         & 100                         & 73                          & 60.7                        & 61.4                        & 64.8                        & 52.8                        & 54.4                        & 62.6                        & 52                          & 54.9                        \\
Annotator 2            & 59.9                        & 48.9                        & 49.8                        & 100                         & 100                         & 100                         & 63.1                        & 54.3                        & 54.9                        & 55.3                        & 43.5                        & 45.4                        \\
Annotator 3            &\cellcolor{orange!40}{65.9}  &\cellcolor{orange!40}{54}    & 55.3                        &\cellcolor{orange!40}{\textbf{79.3}}  &\cellcolor{orange!40}{\textbf{69.1}}  &\cellcolor{orange!40}{\textbf{69.7}}  & 100                         & 100                         & 100                         &\cellcolor{orange!40}{68.4}  &\cellcolor{orange!40}{56.7}  &\cellcolor{orange!40}{59.7} \\
Annotator 4            & 65                          & 54                          &\cellcolor{orange!40}{56.9}  & 70.5                        & 56.3                        & 58.8                        &\cellcolor{orange!40}{70.6} &\cellcolor{orange!40}{58.2}  &\cellcolor{orange!40}{61.6} & 100            & 100                         & 100                         \\ \hline \hline
RL-CB+MMR              & 52.2                        & 32                          &\cellcolor{yellow!40}{36.8}  & 54.8                        & 37.8                        &\cellcolor{yellow!40}{40.9}   & 56.8                        & 36.8                        &\cellcolor{yellow!40}{39.6}  & 53                          & 32.8                        &\cellcolor{yellow!40}{36.6} \\
RL-CB+TextRank         &\cellcolor{yellow!40}{53}    &\cellcolor{yellow!40}{33.7}  & 34                          &\cellcolor{yellow!40}{56.7}  &\cellcolor{yellow!40}{40.4}  & 34.9                        &\cellcolor{yellow!40}{57.3} &\cellcolor{yellow!40}{36.8} & 36.7                        &\cellcolor{yellow!40}{55.4} &\cellcolor{yellow!40}{37.4} & 34.3                        \\ \hline
RL-GRU+MMR             & 51.9                        & 32.1                        & 36.5                        & 52.9                        & 34.6                        & 38.5                        & 53.2                        & 31.4                        & 36.4                        & 51.3                        & 30.2                        & 36.1                        \\
RL-GRU+TextRank        & 51.5                        & 31.6                        & 33                          & 56.7                        & 40                          & 33.4                        & 54.8                        & 32.8                        & 33.9                        & 49.6                        & 27.3                        & 28.6                        \\ \hline \hline
ST                     & 67.3                        & 55.2                        & 57.8                        & 64.6                        & 51.4                        & 52.1                        & 73.6                        & 60.7                        & 62.5                        & 65.2                        & 51.4                        & 55.1                        \\
ST+RdLoss              & 68                          & 57.6                        & 59.7                        & 63.2                        & 51                          & 51.4                        & 75.3                        & 64.8                        & 65.7                        & 66.8                        & 54.8                        & 57.8                        \\ \hline
MT-Shared               & 70.6                        & 60.1                        & 60.5                        & 64.3                        & 52.2                        & 51.9                        & 74.1                        & 61.9                        & 62.8                        & 66.9                        & 54.6                        & 56.8                        \\
MT-Shared+RdLoss         & 70                          & 59.2                        & 59.7                        & \cellcolor{green!30}{65}    &\cellcolor{green!30}{52.3} & 52.3                        &\cellcolor{green!30}{\textbf{77.3}} &\cellcolor{green!30}{\textbf{67.4}} &\cellcolor{green!30}{\textbf{68.3}} &\cellcolor{green!30}{\textbf{68.8}} & 58.1                        & 60                          \\
MT-Hierarchical        & 70.6                        & 59.9                        & 60.9                        & 63.3                        & 50.6                        & 51.6                        & 77.1                        & 66.5                        & 66.5                        & 68.1                        & 57.1                        & 58.2                        \\
MT-Hierarchical+RdLoss &\cellcolor{green!30}{\textbf{71}}   &\cellcolor{green!30}{\textbf{60.5}} &\cellcolor{green!30}{\textbf{63.1}} & 64.4                        & 52.1                        &\cellcolor{green!30}{53.5} & 75.4                        & 64.3                        & 65.6                        & 68.6                        &\cellcolor{green!30}{\textbf{58.3}} &\cellcolor{green!30}{\textbf{62}} \\ \hline
\end{tabular}
\caption{\textbf{ROUGE} scores averaged for decisions in test set and compared to the four expert annotators. Best score for annotators, baseline approaches, and proposed models are highlighted in orange, yellow, and green, respectively.}
\label{tab:rouge_scores}
\end{table*}

In terms of recall score for the binary classification problem, our proposed models demonstrate significant improvement compared to baseline methods, scoring more than twice the scores achieved by the baseline models (Appendix \ref{appendix: additional_results}). But, the recall metric score has limited use in evaluating the performance of the extractive summarization models as legal documents are often verbose, sometimes containing multiple sentences with similar meanings.

We compare the ROUGE scores for our annotators and models in Table \ref{tab:rouge_scores}. Annotator 3 has the highest ROUGE score among all the other annotators. In terms of ROUGE-1 and ROUGE-2, RL-CB+TextRank performs the best, while RL-CB+MMR scores the highest for ROUGE-L. Overall, CatBoost models perform better than GRU-based models. Also, we observe a sharp decline in ROUGE-2 scores compared to ROUGE-1 scores, indicating the baseline models' limited capability to pick the required sentences for the summary. We observe a general improvement in scores for the proposed approaches when adding the implicit redundancy measure, RdLoss, discussed in Section \ref{section:redundancy_loss}. Also, multi-task (MT) models tend to perform better than single-task (ST) models. Overall, MT-Shared+RdLoss and MT-Hierarchical+RdLoss perform the best, scoring higher than three expert annotators. Our models fall short for Annotator 2 as they have annotated more sentences and thus have more extended summaries than ones generated by our trained models.

We also present the values of the parameter $\lambda$ used in \textit{MMR} and \textit{MMR} in Table \ref{tab:summary_overview}. The models that employ the implicit redundancy check (RdLoss) tend to have higher values for $\lambda$ than those with just an explicit redundancy check, indicating that the additional term in the loss function helps the model tackle redundancy better. 

\begin{table*}[htbp]
\centering
\small
\begin{tabular}{lcr|cr|cr|cr}
\hline
\multicolumn{1}{c}{}          & \multicolumn{2}{c|}{Denied}                         & \multicolumn{2}{c|}{Granted}                        & \multicolumn{2}{c|}{Remanded}                       & \multicolumn{2}{c}{Total}                           \\ \hline
\multicolumn{1}{c}{\textbf{}} & \multicolumn{1}{c}{Rank} & \multicolumn{1}{c|}{Ad.} & \multicolumn{1}{c}{Rank} & \multicolumn{1}{c|}{Ad.} & \multicolumn{1}{c}{Rank} & \multicolumn{1}{c|}{Ad.} & \multicolumn{1}{c}{Rank} & \multicolumn{1}{c}{Ad.} \\ \hline \hline
Annotator 3                   & \cellcolor{orange!40}{2.83 ± 1.17}             & \cellcolor{orange!40}{0.67}              & 3.75 ± 1.89           & 0.50              & \cellcolor{orange!40}{2.9 ± 1.37}            & \cellcolor{orange!40}{\textbf{1.00}}       & \cellcolor{orange!40}{3.05 ± 1.39}                     & \cellcolor{orange!40}{\textbf{0.8}}  \\
Annotator 4                   & 4.17 ± 0.75             & 0.50              & \cellcolor{orange!40}{2.5 ± 1.0}             & \cellcolor{orange!40}{\textbf{0.75}}     & 2.9 ± 1.73            & 0.90                & 3.2 ± 1.47                      & 0.75 \\ \hline
MT-Hierarchical+RdLoss        & \cellcolor{green!30}{\textbf{1.83 ± 1.6}}     & \cellcolor{green!30}{\textbf{0.83}}     & 2.5 ± 1.91            & \cellcolor{green!30}{\textbf{0.75}}     & 3.3 ± 1.57            & 0.60                & 2.7 ± 1.69                      & 0.7  \\
MT-Shared+RdLoss              & 2.0 ± 1.26              & \textbf{0.83}     & 2.25 ± 1.89           & 0.50              & 2.3 ± 1.42            &\cellcolor{green!30}{0.80}                & 2.2 ± 1.4                       & \cellcolor{green!30}{0.75} \\
ST+RdLoss                     & 2.5 ± 1.64              & 0.67              & \cellcolor{green!30}{\textbf{2.0 ± 1.41}}   & \cellcolor{green!30}{\textbf{0.75}}     &\cellcolor{green!30}{\textbf{ 1.8 ± 1.03}}   & \cellcolor{green!30}{0.80}                & \cellcolor{green!30}{\textbf{2.05 ± 1.28 }}           & \cellcolor{green!30}{0.75} \\ \hline
\end{tabular}
\caption{Qualitative analysis of the summaries in terms of \textbf{Rank} (lower is better) and \textbf{Adequacy} (higher is better). Best score for annotator and proposed models are highlighted in orange and green, respectively.}
\label{tab:qualitative}
\end{table*}

\begin{figure}[htbp!]
    \centering
    \includegraphics[width=1\linewidth]{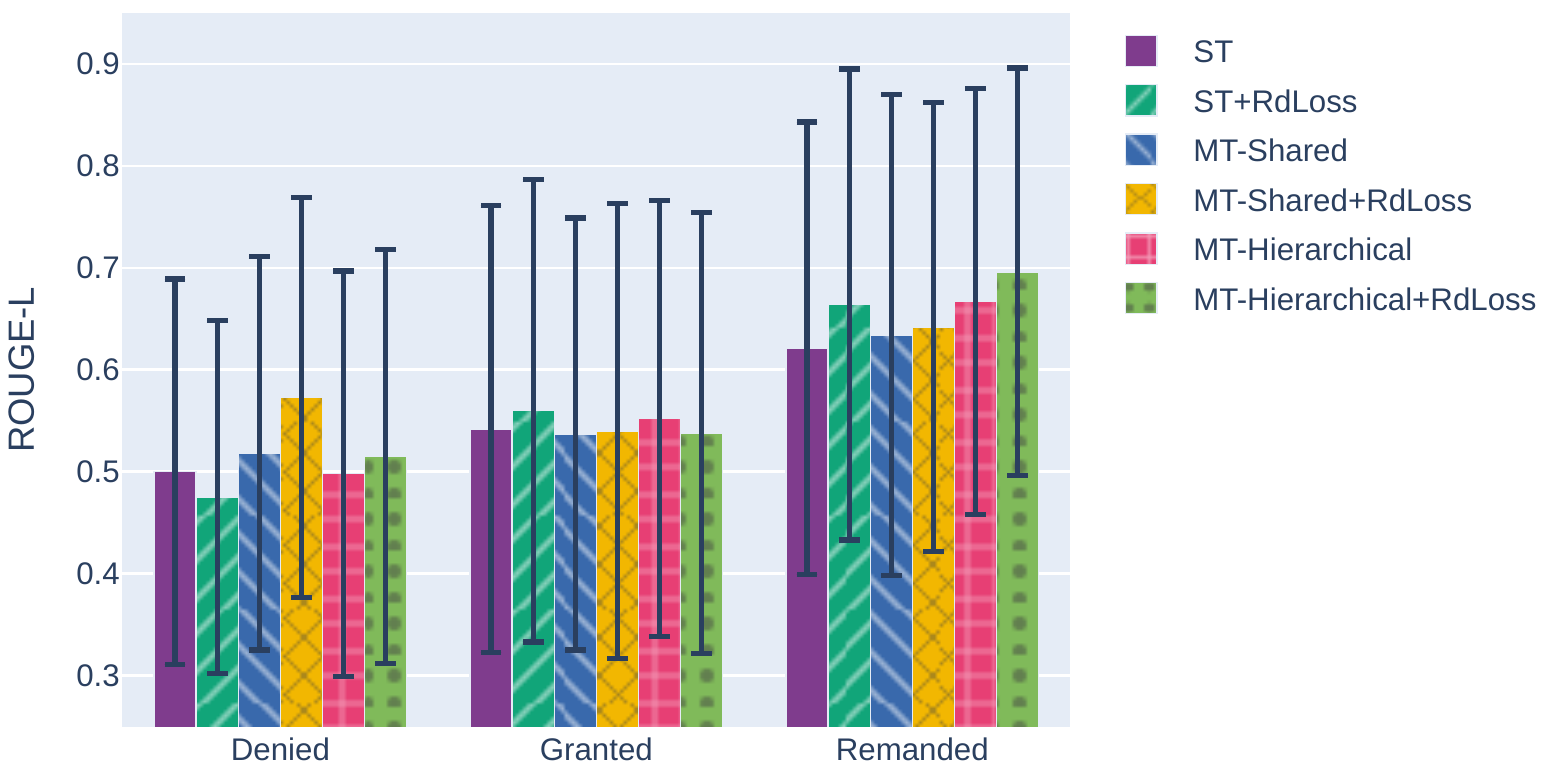}
    \caption{Comparison of the performance for different decision outcomes in terms of ROUGE-L.}
    \label{fig:model_comparisons_outcome}
\end{figure}

\begin{figure}[htbp!]
    \centering
    \includegraphics[width=1\linewidth]{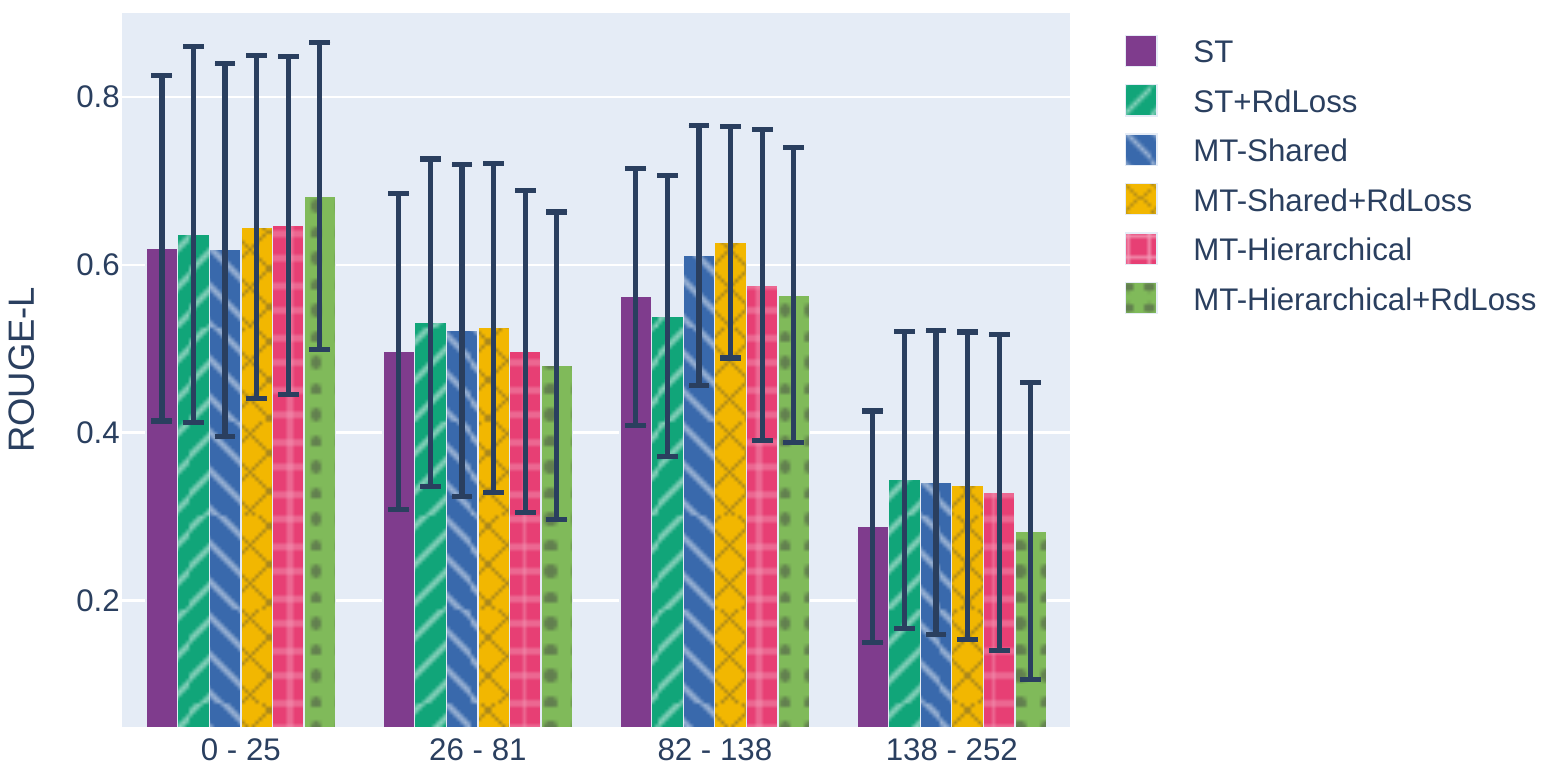}
    \caption{Performance for different methods based on the number of sentences in the decisions in terms of ROUGE-L.}
    \label{fig:model_comparisons_length}
\end{figure}

We also compare the performance in terms of the outcome of the legal decisions in Figure \ref{fig:model_comparisons_outcome}. The findings with the remanded outcome have higher scores, constituting approximately 50\% of the train and test data. MT-Shared+RdLoss performs notably better for denied outcomes, while MT-Hierarchical+RdLoss scores the highest for remanded decisions. The shared layers between the primary and auxiliary tasks combined with a supplementary dataset consisting of only granted and denied findings make it difficult for the MT-Shared and MT-Shared+RdLoss to generate better summaries for remand cases. Also, remanded cases can differ in terms of rhetorical role distribution from granted and denied decisions, as they end in the BVA sending the case back to the regional VA office, often instructing it extensively on what to do next. In contrast, the hierarchical multi-task (MT) models leverage the additional GRU layer to perform better for such outcomes. As shown in Figure \ref{fig:model_comparisons_length}, we note a significant decrease in performance for lengthy decisions for all the methods that could be attributed both to the limitations of the GRU models and insufficient training data for such cases.

\subsection{Qualitative Evaluation}

We further perform manual qualitative analysis to better understand the proposed approaches' performance compared to the expert annotators. We generated the summaries for each decision in the test set using three of the proposed models and compared them with the ones from two expert annotators. The outputs were randomized, and a fifth annotator with expertise in the legal domain (the third author) ranked each summary and checked if it was adequate against the two human-drafted reference summaries. Following \cite{zhong2019automatic}, we consider a summary adequate if it identifies all major legal issues and resolutions in the case. We then rank summaries based on the additional information they contain about the case narrative and their coherence. We assign the same rank to two or more summaries if they are duplicate, near-duplicate, or semantically equivalent.

We report the results for qualitative analysis in Table \ref{tab:qualitative}. Overall our proposed models almost always ranked better than the two annotators. ST+RdLoss ranked best on average and for decisions with the outcome as remanded or granted. MT-Hierarchical+RdLoss ranked better for the denied decisions. In terms of adequacy, both Annotator 3 and MT-Hierarchical+RdLoss achieved reasonable accuracy. The multi-task (MT) models achieve higher ROUGE scores but fail to produce summaries of good qualities for remand cases. A possible explanation is that the design of the annotation type system is more suitable for evidence-based findings (e.g., an in-service stressor has caused a particular disability of the veteran), which are most clearly present in denied and granted cases. Remand cases challenge this annotation type system because the BVA determines that we cannot make a finding based on the current evidence. Summary annotators must cope with this slight semantic mismatch and may develop an individual lexical bias in assigning sentence types. The MT-Hierarchical+RdLoss model can potentially overfit that bias because it contains an additional GRU layer unaffected by the rhetorical role supervision signal.

\section{Discussion}

Our proposed approaches generate higher-scoring summaries than baseline methods and expert annotators in terms of ROUGE scores and appear to be competitive in a qualitative expert ranking evaluation. The domain-specific pre-trained transformers help produce sentence embeddings capable of capturing the semantics of legal decisions in a way conducive to being used as a component in our proposed setup for summary generation. We can use these embeddings to train simple models like GRU effectively. Adding explicit methods like \textit{MMR} helps tackle redundancy to generate more informative summaries even for verbose legal decisions. We further improve the performance by using the weighted cross-entropy loss function combined with redundancy loss (\textit{RdLoss}). The additional loss term helps train models that can handle redundancy implicit.

The quantitative measures show the effectiveness of both the single-task and multi-task models, even with a limited dataset containing just 120 legal decisions. The supplementary dataset used to train the same model for the rhetorical role labeling task helps learn better representations and achieve better summary ROUGE scores. Such correlated tasks and datasets prove helpful in low-resource settings such as ours to improve the performance further at no additional annotation costs, assuming they stem from the same legal domain. Further qualitative analysis with the help of an expert annotator shows our proposed approaches rank at least as good as human annotators. It also indicates the need for better quantitative metrics to evaluate the quality of the summaries. The specialized BVA domain we experiment in is relatively narrow in scope and highly regular in its document structure. Our proposed methods seem promising as they work well with limited annotated datasets and computational resources but warrant further investigation and validation on larger, more diverse datasets.

\section{Conclusion}
Training models that can generate extractive summaries of legal opinions comparable to expert annotators in a low-resource setting can be challenging. We demonstrate that  domain-specific pre-trained transformers and multi-task training with rhetorical role labeling can effectively train sequential extractive summarizaton models (in our case, GRUs) on a relatively constrained domain of cases. The proposed methods implement implicit and explicit redundancy checks to maximize the information and minimize the redundancy in summaries. In our experiments, we systematically analyze the performance of different techniques, both quantitatively and qualitatively. The results verify the efficacy of our model design. In the future, we plan to extend our work to other decision types and jurisdictions in the legal domain. We further plan to explore the discrepancy between our quantitative results favoring multi-task models and qualitative evaluation preferring summaries by single-task architectures.
    
\section*{Limitations}
We only consider two different rhetorical roles for summarization. Decisions from other subdomains and jurisdictions might require the inclusion of more rhetorical functions in the final summary. A suitable domain-specific pre-trained transformer might not be available to produce the necessary sentence embeddings. In such cases, we would have to rely on conventional approaches like Universal Sentence Encoder (USE) \citep{cer-etal-2018-universal}, GloVe \citep{pennington-etal-2014-glove}, and Word2Vec \citep{NIPS2013_9aa42b31}. Our methods do not automatically scale well for very long decisions, so we must ensure ample availability of such decisions in the training set.

The BVA decisions we use have a relatively regular structure and are constrained to cases deciding issues of compensation for service-connected PTSD disabilities of veterans, which is only a subset of the issues adjudicated by the BVA. The decisions discuss similar aspects of medical diagnoses, stressors experienced in service, and causation. We still need to validate if we could extend our proposed approach to collections of cases that include more diverse legal issues and fact patterns. Legal texts often contain language that looks similar on the surface but is different in its semantics and vice versa. More complex textual phenomena may challenge the redundancy-focused components of the system.

\section*{Ethical Concerns}

The Board of Veterans' Appeals publicly releases its decisions (including the ones in our datasets) on its publicly available website. Generally, in BVA decisions, the veteran is not named explicitly. While the nature of disability compensation claims (including those relating to PTSD) is sensitive, we chose these particular datasets because of several factors: (a) the scarcity of expert-annotated legal decisions in the public domain was suitable for summarization research when conducting the experiments; (b) the availability of PTSD-related annotated decisions from the Research Laboratory for Law, Logic, and Technology (LLT Lab) at Hofstra University with a matching set of summaries and (c) our prior experience where we worked with BVA decisions and U.S. Veterans Law.
Additionally, proving the requirements of a service-connected PTSD disability using relevant evidence is legally sufficiently complex to provide a suitable testbed to evaluate the proposed summarization techniques. At the same time, it is a reasonably closed domain such that the comparative ranking of candidate summaries is more straightforward and coherent.

The biases, inequalities, and under-representations encoded in the pre-trained transformer models might get inherited by our GRU models and propagated to the generated summaries  \citep{bommasani-etal-2020-interpreting}. To deploy these models in a production system, one must thoroughly check for such biases by comprehensively evaluating summarization performance across relevant groups (e.g., gender and race) using tests such as, for example, the recently proposed  Pronoun-Ranking Test \citep{silva2021towards}.

An automatic summarization model for legal decisions may perform worse for some partitions of its domain than others. For example, in the BVA context, cases about rarely occurring disabilities or special legal and military situations may lead to worse summaries because of sparseness in the training data. It might disparately affect groups that are supposed to be treated equally if group membership tends to coincide with such configurations. If lawyers use (and potentially depend on) automatic summarization tools to assist clients, screening such systems may become necessary. For example, one can engage domain experts to curate datasets with better representation across different types of injuries and legal phenomena that might be uncommon or related to particular groups. Still, the quantitative improvement of additional development data obtained toward more consistent summary quality may be uncertain and model-dependent. At the very least, it will help reveal performance disparities and increase expert user awareness around the limitations of automatic summarization technology in the legal domain. 

\section{Acknowledgements}
We want to thank Rashid Haddad and Santosh T.Y.S.S. from the Technical University of Munich for their assistance in analyzing the results and feedback about the manuscript. 

\bibliography{anthology,custom}
\bibliographystyle{acl_natbib}

\newpage
\appendix

\section{Inter-Annotator Agreement}
\label{sec:appendix_cohen}

We present the pairwise inter-annotator agreement score using the Cohen-Kappa coefficient in Table \ref{tab:cohen_kappa_score_annotators}. The scores vary from 0.46 to 0.55, indicating low agreement among annotators. However, this metric is not ideal, as annotators can mark up different sentences that still address similar aspects of the case. We also calculate the pairwise metric for our proposed methods in Table \ref{tab:cohen_kappa_score_model} to measure how similar the outputs are. The highest agreement is between MT-Shared+RdLoss and MT-Hierarhical+RdLoss. The summaries generated by these two approaches mostly differ by 1 or 2 sentences. 

\begin{table*}[htbp]
\centering
\small
\begin{tabular}{lrrrr}
\hline
\multicolumn{1}{c}{} & \multicolumn{1}{c}{Annotator 1} & \multicolumn{1}{c}{Annotator 2} & \multicolumn{1}{c}{Annotator 3} & \multicolumn{1}{c}{Annotator 4} \\ \hline \hline
Annotator 1          & 100                               & \textbf{49.4}                  & 46.1                           & 46.7                           \\
Annotator 2          & 49.4                           & 100                               & \textbf{54.4}                  & 46.7                           \\
Annotator 3          & 46.1                           & \textbf{54.4}                  & 100                               & 49.4                           \\
Annotator 4          & 46.7                           & 46.7                           & \textbf{49.4}                  & 100                               \\ \hline
\end{tabular}
\caption{Cohen Kappa Score for the four annotators}
\label{tab:cohen_kappa_score_annotators}

\bigskip

\centering
\small
\begin{tabular}{lcc|cccc}
\hline
\multicolumn{1}{c}{} & \multicolumn{1}{c}{ST} & \multicolumn{1}{c|}{ST+RdLoss} & \multicolumn{1}{c}{MT-Shared} & \multicolumn{1}{c}{MT-Shared+RdLoss} & \multicolumn{1}{c}{MT-Hier} & \multicolumn{1}{c}{MT-Hier+RdLoss} \\ \hline \hline
ST                   & 100                  & 73.5                      & \textbf{74.4}                & 70.2                                & 65.7                       & 63.8                              \\
ST+RdLoss            & 73.5                 & 100                        & \textbf{75.5}             & 73.3                                & 71.2                       & 72.8                              \\ \hline
MT-Shared            & 74.4                 & 75.5                       & 100                          & \textbf{79.7}                       & 73.3                       & 72.3                              \\
MT-Shared+RdLoss     & 70.2                  & 73.3                         & 79.7                         & 100                                    & 77.9                       & \textbf{81.5}                     \\ 
MT-Hier              & 65.7                  & 71.2                         & 73.3              & 77.9                                & 100                           & \textbf{81.2}                     \\
MT-Hier+RdLoss              & 63.8                  & 72.8                         & 72.3                & \textbf{81.5}                       & 81.2                       & 100                \\      \hline           
\end{tabular}
\caption{Cohen Kappa Score for the six proposed methods}
\label{tab:cohen_kappa_score_model}

\label{tab:quality_overview}

\end{table*}

\section{Sentence Embeddings Visualization}

\begin{figure}[htbp!]
    \centering
    \includegraphics[width=0.95\linewidth]{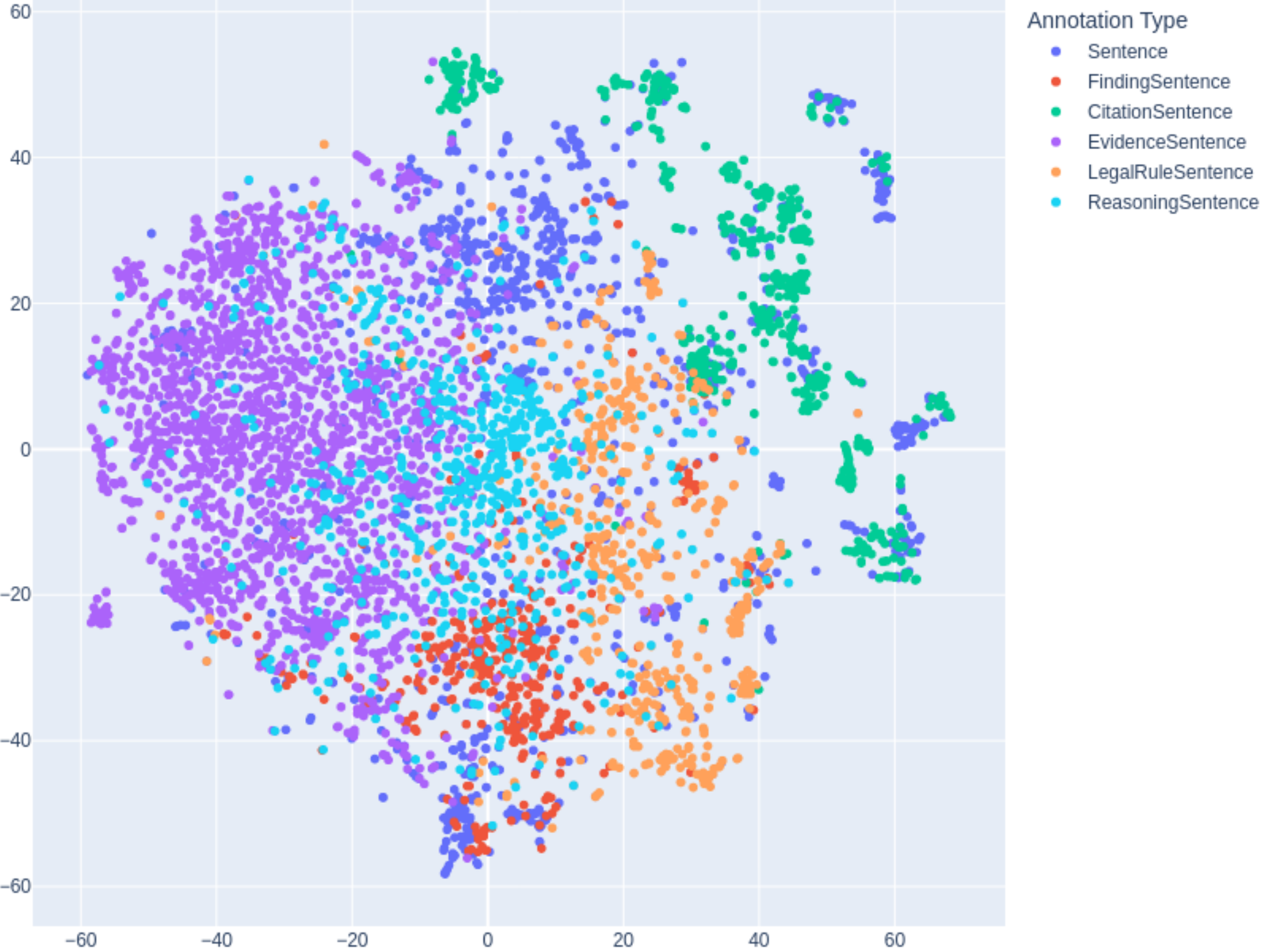}
    \caption{T-SNE projection of the 768-dimensional sentence embeddings to two-dimensional space for the 6984 sentences and corresponding rhetorical roles.}
    \label{fig:rhetorical_embeddings}
\end{figure}

We illustrate the T-SNE projection of the sentence embeddings for the 50 decisions in the rhetorical labeling dataset in Figure \ref{fig:rhetorical_embeddings}. The sentence embeddings can easily separate annotation types like Citation, LegalRule, and Finding sentences, as these classes have a somewhat unique vocabulary. However, the embeddings for other annotation types do not have such a clear distinction and overlap. 

\section{Additional Results}\label{appendix: additional_results}

For the Recall score presented in Table \ref{tab:extractive_recall}, the baseline model, RL-CB+TextRank, performs the best for three of the four annotators, while RL- CB+MMR scores the highest for the remaining annotator. Our proposed models perform notably better than the baseline models. Overall, the multi-task (MT) based models outperform the single-task (ST) models. MT-Hierarchical+RdLoss performs the best for Annotator 1 and Annotator 4.

\begin{table*}[htbp!]
\centering
\small
\begin{tabular}{lr|r|r|r}
\hline
\multicolumn{1}{c}{}     & \multicolumn{1}{c|}{Annotator 1} & \multicolumn{1}{c|}{Annotator 2} & \multicolumn{1}{c|}{Annotator 3} & \multicolumn{1}{c}{Annotator 4} \\ \hline \hline
RL-CB+MMR              & 0.255 ± 0.265                        & 0.308 ± 0.272                        & \cellcolor{yellow!40}{0.283 ± 0.261} & 0.231 ± 0.263                        \\
RL-CB+TextRank         & \cellcolor{yellow!40}{0.285 ± 0.29}  & \cellcolor{yellow!40}{0.35 ± 0.247}  & 0.276 ± 0.25                         & \cellcolor{yellow!40}{0.294 ± 0.24}  \\ \hline
RL-GRU+MMR             & 0.227 ± 0.218                        & 0.25 ± 0.22                          & 0.215 ± 0.236                        & 0.191 ± 0.273                        \\
RL-GRU+TextRank        & 0.243 ± 0.282                        & 0.333 ± 0.286                        & 0.247 ± 0.279                        & 0.18 ± 0.248                         \\ \hline \hline
ST                     & 0.476 ± 0.257                        & 0.45 ± 0.254                         & 0.52 ± 0.25                          & 0.422 ± 0.263                        \\
ST+RdLoss              & 0.499 ± 0.267                        & 0.467 ± 0.251                        & 0.578 ± 0.233                        & 0.457 ± 0.273                        \\ \hline
MT-Shared              & 0.498 ± 0.242                        & \cellcolor{green!30}{\textbf{0.475 ± 0.225}}  & 0.528 ± 0.238                        & 0.441 ± 0.237                        \\
MT-Shared+RdLoss       & 0.486 ± 0.238                        & 0.45 ± 0.23                          & 0.58 ± 0.236                         & 0.466 ± 0.237                        \\
MT-Hierarchical        & 0.508 ± 0.272                        & 0.458 ± 0.259                        & \cellcolor{green!30}{\textbf{0.584 ± 0.255}} & 0.467 ± 0.254                        \\
MT-Hierarchical+RdLoss & \cellcolor{green!30}{\textbf{0.519 ± 0.274}}  & 0.467 ± 0.251                        & 0.544 ± 0.239                        & \cellcolor{green!30}{\textbf{0.495 ± 0.265}}  \\ \hline
\end{tabular}
\caption{\textbf{Recall} scores for the extractive summarization task averaged for all the decisions in the test set. Best score for baseline and proposed models are highlighted in yellow and green, respectively.}
\label{tab:extractive_recall}
\end{table*}

\section{Hyperparameter Tuning for Baseline Methods}\label{appendix:baseline}

\begin{figure*}[htbp!]
    \centering
    \includegraphics[width=0.95\linewidth]{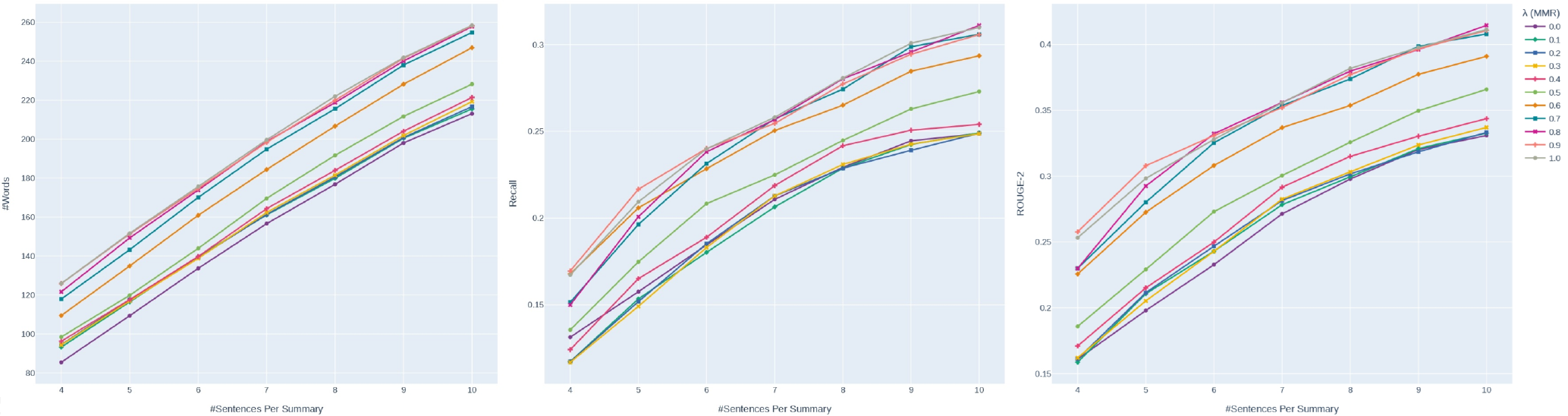}
    \caption{We estimate the best value for the parameter $\lambda$ required for \textit{MMR} by varying it from 0 to 1 at intervals of 0.1 for the training set.}
    \label{fig:catboost_mmr_lambda}
\end{figure*}

This section briefly discusses how we determine the parameter $\lambda$ for \textit{MMR} in the baseline methods: RL-CB+MMR and RL-GRU+MMR. Once we have filtered out the Evidence/Reasoning sentences using the CatBoost or GRU classifier, we generate the summaries using different values of $\lambda$ for all the decisions in the training set. We also vary the number of sentences to determine the optimal length for the summary. We then measure the impact of $\lambda$ and summary length on the total number of words in the summary, recall score, and ROUGE-L, as shown in Figure \ref{fig:catboost_mmr_lambda}. We choose values that result in the highest recall and ROUGE-L but consist of the number of tokens comparable to experts and proposed methods. The same values are used to produce the summaries for decisions in the test set. 

\section{Hyperparameters tuning for Proposed Methods}\label{appendix:proposed_models}

We use a combination of random and bayesian searches to find the best set of hyperparameters for our models. We use the random search to find an approximate search space suitable for our model, followed by a more targeted search using bayesian optimization. All our models achieve the best scores using just one GRU layer. Also, since we use sentence embeddings derived from pre-trained transformers, our models converge quickly with a minimal number of epochs. We report the best set of hyperparameters for all our models in Table \ref{tab:proposed_models_hyperparameter}

\begin{table*}[htbp!]
\centering
\small
\begin{tabular}{lcc|cc|cc|cc}
\hline
\multicolumn{1}{c}{} & \multicolumn{1}{c}{num\_layers} & \multicolumn{1}{c|}{hidden\_size} & \multicolumn{1}{c}{dropout (RL)} & \multicolumn{1}{c|}{dropout (ES)} & \multicolumn{1}{c}{batch\_size} & \multicolumn{1}{c|}{epochs} & \multicolumn{1}{c}{learning\_rate} \\ \hline 
ST                   & 1                               & 128                              & -                                & 0.5                              & 8                               & 5                          & 0.00261                \\
ST+RdLoss            & 1                               & 64                               & -                                & 0.6                              & 8                               & 9                          & 0.00441                \\ \hline
MT-Shared            & 1                               & 512                              & 0.4                              & 0.5                              & 4                               & 6                          & 0.00019                \\
MT-Shared            & 1                               & 512                              & 0.6                              & 0.4                              & 4                               & 11                         & 0.00018                \\ \hline
MT-Hier              & 1 + 1                            & 128 + 512                        & 0.5                              & 0.6                              & 8                               & 5                          & 0.00143                \\
MT-Hier+RdLoss       & 1 + 1                           & 128 + 256                        & 0.6                              & 0.4                              & 4                               & 8                          & 0.00053               \\ \hline
\end{tabular}
\caption{Final hyperparameters required to train the proposed extractive summarization models.}
\label{tab:proposed_models_hyperparameter}
\end{table*}

\begin{figure*}[htbp!]
    \centering
    \includegraphics[width=0.85\linewidth]{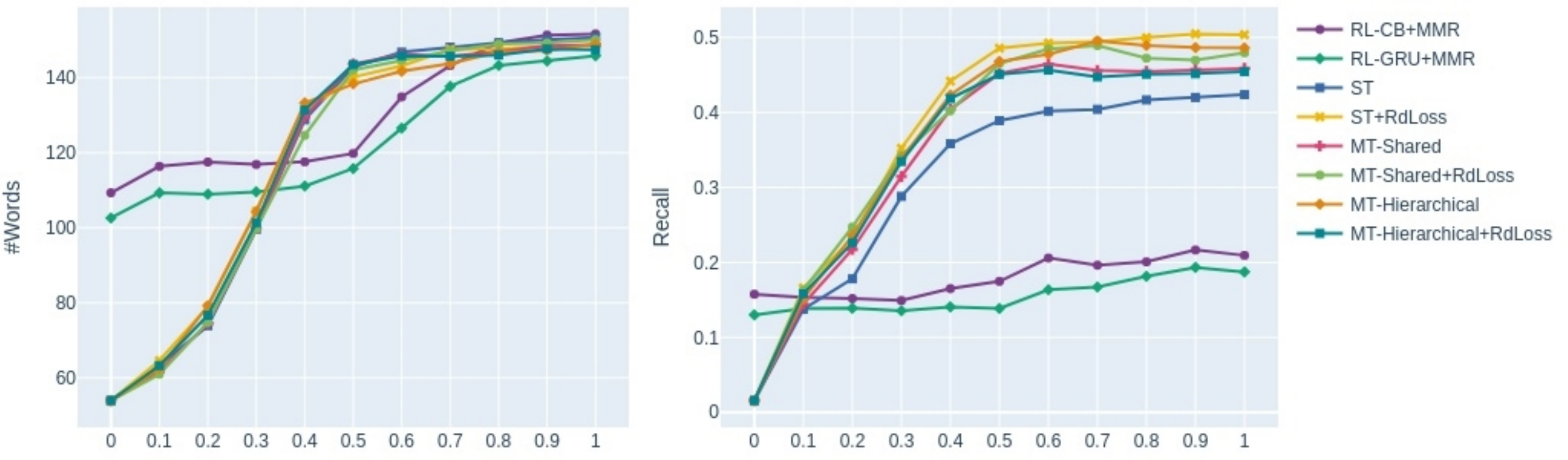}
    \caption{Effect of varying $\lambda$ for different models on training set in terms of number of words in summary, extractive Recall and ROUGE-L scores}
    \label{fig:lambda_effect}
\end{figure*}

In Figure \ref{fig:lambda_effect}, we demonstrate the effect of $\lambda$ on our proposed methods for the training set. The number of tokens and recall score increase linearly with the value of $\lambda$ up to 0.5 and then changes very slowly. Assigning less weight to $\lambda$ forces the model to pick very dissimilar sentences and results in poor performance. Therefore, we set the minimum value of $\lambda$ as 0.5 for hyperparameter tuning to find the right balance between similarity and redundancy. Additionally, the $\lambda$ has a minimal impact on baseline methods.

\section{Examples}
Based on the assessment of our qualitative analysis, we demonstrate the summaries generated by annotators and models for two decisions, one each from denied and remanded outcomes (ordered by their rank). The multi-task models generate better summaries than the two annotators in the example shown in \ref{appendix:denied_example}. For the second example in \ref{appendix:remand_example}, the annotators are ranked better than all the proposed methods, which fail to produce adequate summaries. 

\subsection{Denied Example}\label{appendix:denied_example}

\noindent\textbf{\textit{MT-Hierarchical+RdLoss:}}

\noindent \textit{``Furthermore, Vietnam casualty logs do not confirm the supposed death of PL nor do the deck logs record PLs unavailability for duty due to injury or death. Overall, the veterans report of stressors while serving on the USS CORAL SEA lack any indicia of reliability or credibility. Certainly, there is no credible supporting evidence of these alleged stressors. Based upon the above, the Board finds that the preponderance of the evidence establishes that the veteran was not engaged in combat, and that his allegations of exposure to combat and non-combat stressors are not credible. The Board, therefore, finds that the veterans PTSD is not shown to have its origins in his military service.''}

\noindent\textbf{\textit{MT-Shared+RdLoss:}}

\noindent \textit{``The Board also notes that the veteran appears to have recanted his original assertion, contained in a March 1995 VA mental health clinic visit, that he "frequently ran into combat" on mainland visits. Overall, the veterans report of stressors while serving on the USS CORAL SEA lack any indicia of reliability or credibility. Certainly, there is no credible supporting evidence of these alleged stressors. Based upon the above, the Board finds that the preponderance of the evidence establishes that the veteran was not engaged in combat, and that his allegations of exposure to combat and non-combat stressors are not credible. The Board, therefore, finds that the veterans PTSD is not shown to have its origins in his military service.''}

\noindent\textbf{\textit{Annotator 4:}}

\noindent \textit{``The veteran's service medical records do not reflect his report of exposure to combat or non-combat stressors. Overall, the veteran's report of stressors while serving on the USS CORAL SEA lack any indicia of reliability or credibility. Based upon the above, the Board finds that the preponderance of the evidence establishes that the veteran was not engaged in combat, and that his allegations of exposure to combat and non-combat stressors are not credible. The Board, therefore, finds that the veteran's PTSD is not shown to have its origins in his military service.''}

\noindent\textbf{\textit{Annotator 3:}}

\noindent \textit{``The veteran's service medical records do not reflect his report of exposure to combat or non-combat stressors. There is no report that an individual named PL was unavailable for duty due to injury or death. The Board, therefore, finds that the veteran's PTSD is not shown to have its origins in his military service. The claim for service connection for PTSD, therefore, must be denied.''}

\noindent\textbf{\textit{ST+RdLoss:}}

\noindent \textit{``Overall, the veterans report of stressors while serving on the USS CORAL SEA lack any indicia of reliability or credibility. Certainly, there is no credible supporting evidence of these alleged stressors. Based upon the above, the Board finds that the preponderance of the evidence establishes that the veteran was not engaged in combat, and that his allegations of exposure to combat and non-combat stressors are not credible. The Board, therefore, finds that the veterans PTSD is not shown to have its origins in his military service. As the veteran was not engaged in combat, the evidentiary presumptions contained in 38 U.S.C.A. § 1154(b) and 38 C.F.R. § 3.304(f) are not applicable in this case.''}

\subsection{Remand Example}\label{appendix:remand_example}

\noindent\textbf{\textit{Annotator 4:}}

\noindent \textit{``The remand instructions stated that the examiner was to render an opinion as to whether the documented in-service headaches and fainting spells were manifestations of her diagnosed acquired psychiatric disorder (to include PTSD, depression, and anxiety). Regarding the Veterans complaints of headaches and fainting spells in service, the examiner stated, "The Veteran reports multiple pain issues including headaches and back pain. This, however, does not respond to the question as to whether these headaches and fainting spells were manifestations of a current psychiatric disorder. In addition, the Veteran, through her representative, has asserted that her PTSD and depression are caused or aggravated by her service connected migraines. As the December 2016 opinion does not clearly address this issue, an additional medical opinion is warranted to determine whether the Veterans PTSD is proximately due to or aggravated by her service-connected migraines.''}

\noindent\textbf{\textit{Annotator 3:}}

\noindent \textit{``The examiner noted mental health diagnoses of chronic PTSD with secondary generalized anxiety disorder and major depressive disorder. Regarding the Veterans complaints of headaches and fainting spells in service, the examiner stated, "The Veteran reports multiple pain issues including headaches and back pain. This, however, does not respond to the question as to whether these headaches and fainting spells were manifestations of a current psychiatric disorder. In addition, the Veteran, through her representative, has asserted that her PTSD and depression are caused or aggravated by her service connected migraines. The examiner must also opine as to whether it is at least as likely as not that the in-service episodes of fainting and headaches were a manifestation of a currently diagnosed acquired psychiatric disorder.''}

\noindent\textbf{\textit{MT-Shared+RdLoss:}}

\noindent \textit{``The remand instructions stated that the examiner was to render an opinion as to whether the documented in-service headaches and fainting spells were manifestations of her diagnosed acquired psychiatric disorder (to include PTSD, depression, and anxiety). The examiner was also advised to address the medical literature in the October 2016 Appellate Brief suggesting that fainting and headaches can be physical symptoms of PTSD and can occur as a result of exposure to trauma and the Veterans September 1980 in-service reports of headaches and fainting spells. Regarding the Veterans complaints of headaches and fainting spells in service, the examiner stated, "The Veteran reports multiple pain issues including headaches and back pain. As the December 2016 opinion does not clearly address this issue, an additional medical opinion is warranted to determine whether the Veterans PTSD is proximately due to or aggravated by her service-connected migraines. The examiner must also opine as to whether it is at least as likely as not that the in-service episodes of fainting and headaches were a manifestation of a currently diagnosed acquired psychiatric disorder.''}

\noindent\textbf{\textit{ST+RdLoss:}}

\noindent \textit{``The Veteran was afforded an additional VA psychiatric evaluation in December 2016.  The examiner noted mental health diagnoses of chronic PTSD with secondary generalized anxiety disorder and major depressive disorder.  Regarding the Veteran's complaints of headaches and fainting spells in service, the examiner stated, "The Veteran reports multiple pain issues including headaches and back pain.  These issues could be exacerbated by emotional distress but are not directly related."  This, however, does not respond to the question as to whether these headaches and fainting spells were manifestations of a current psychiatric disorder. This statement requires clarification. In addition, the Veteran, through her representative, has asserted that her PTSD and depression are caused or aggravated by her service connected migraines.  See June 2017 Appellate Brief.  As the December 2016 opinion does not clearly address this issue, an additional medical opinion is warranted to determine whether the Veteran's PTSD is proximately due to or aggravated by her service-connected migraines.''}

\bigskip

\noindent\textbf{\textit{MT-Hierarchical+RdLoss:}}

\noindent \textit{`` The Veteran was afforded an additional VA psychiatric evaluation in December 2016. This, however, does not respond to the question as to whether these headaches and fainting spells were manifestations of a current psychiatric disorder. In addition, the Veteran, through her representative, has asserted that her PTSD and depression are caused or aggravated by her service connected migraines. See June 2017 Appellate Brief. As the December 2016 opinion does not clearly address this issue, an additional medical opinion is warranted to determine whether the Veteran's PTSD is proximately due to or aggravated by her service-connected migraines. Obtain a VA addendum opinion by the same December 2016 examiner, (or another appropriate examiner if unavailable), to provide opinions as to whether it is at least as likely as not (50 percent or better probability) that the Veteran's PTSD was caused by OR aggravated (i.e., permanently worsened beyond the natural progress of the disorder) by her service-connected migraine headaches.''}

\end{document}